\documentclass{article}

\usepackage{arxiv}

\usepackage[utf8]{inputenc} 
\usepackage[T1]{fontenc}    
\usepackage{hyperref}       
\usepackage{url}            
\usepackage{booktabs}       
\usepackage{amsfonts}       
\usepackage{nicefrac}       
\usepackage[final,nopatch=footnote]{microtype}      
\usepackage{lipsum}		
\usepackage{graphicx}
\usepackage{natbib}
\usepackage{doi}

\usepackage{amsmath}
\usepackage{svg}
\usepackage{subcaption}
\usepackage{linguex}

\title{SNAPE-PM: Building and Utilizing Dynamic Partner Models for
Adaptive Explanation Generation}

\author{ \href{https://orcid.org/0000-0001-5622-8248}{\includegraphics[scale=0.06]{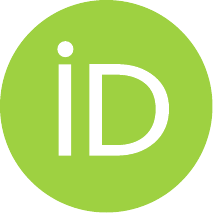}\hspace{1mm}Amelie S. Robrecht} \\
	Social Cognitive Systems\\
    TRR 318 | Constructing Explainability\\
	Bielefeld University\\
	Germany \\
    \And
	\href{https://orcid.org/0009-0009-4668-6785}{\includegraphics[scale=0.06]{orcid}\hspace{1mm}Christoph R. Kowalski} \\
	Social Cognitive Systems\\
    TRR 318 | Constructing Explainability\\
	Bielefeld University\\
	Germany \\
	\And
	\href{https://orcid.org/0000-0002-4047-9277}{\includegraphics[scale=0.06]{orcid}\hspace{1mm}Stefan Kopp} \\
	Social Cognitive Systems\\
    TRR 318 | Constructing Explainability\\
	Bielefeld University\\
	Germany \\
}

\renewcommand{\shorttitle}{SNAPE-PM}

\hypersetup{
pdftitle={SNAPE-PM},
pdfsubject={q-bio.NC, q-bio.QM},
pdfauthor={Amelie S. Robrecht, Christoph R. Kowalski, Stefan Kopp},
pdfkeywords={Explainability, Adaptivity, Partner Models, Dynamic Bayesian Network, Non-stationary Decision Process},
}

\begin{document}
\maketitle

\begin{abstract}
	Adapting to the addressee is crucial for successful explanations, yet poses significant challenges for dialogsystems. We adopt the approach of treating explanation generation as a non-stationary decision process, where the optimal strategy varies according to changing beliefs about the explainee and the interaction context. In this paper we address the questions of (1) how to track the interaction context and the relevant listener features in a formally defined computational partner model, and (2) how to utilize this model in the dynamically adjusted, rational decision process that determines the currently best explanation strategy. We propose a Bayesian inference-based approach to continuously update the partner model based on user feedback, and a non-stationary Markov Decision Process to adjust decision-making based on the partner model values. We evaluate an implementation of this framework with five simulated interlocutors, demonstrating its effectiveness in adapting to different partners with constant and even changing feedback behavior. The results show high adaptivity with distinct explanation strategies emerging for different partners, highlighting the potential of our approach to improve explainable AI systems and dialogsystems in general.
\end{abstract}


\section{Introduction}
When giving an explanation, people convey ideas, beliefs, objects, or processes to one another. 
In literature, an explanation is either seen as a process (an interaction involving at least two agents) or as a product (an answer to a why question) \citep{lombrozo_structure_2006}. Explanations thereby can address three types of questions: What?, How?, and Why? \citep{miller_explanation_2019}. Most approaches to making AI systems explainable have focused on explanations as single-turn answers to why-questions \citep{chandra_cooperative_2024, lewis_causal_1986, anjomshoae_explainable_2019}. However, explanations must also be considered as being constructed through a collaborative interaction process involving at least two agents  \citep{rohlfing_explanation_2021,robrecht_snape_2023} with the potential to address each type of explanatory question (cf.~\citep{axelsson_using_2023,el-assady_towards_2019}). In this view, an agent with advanced knowledge (the explainer) tries to tailor the explanation to an agent with less knowledge (the explainee), who in turn provides feedback and thus contributes to the joint co-construction of the explanation. The ermerging interaction is influenced by the explainer's actions and the explainee's feedback, their respective roles, or the domain being discussed. Although the advantage of interactive explanations is also known in the XAI community, most research in the field provides static explanations \cite{ali_explainable_2023} or accepts major limitations in terms of dialog planning and informativeness \citep{piriyakulkij_active_2023, bertrand_selective_2023}.

Crucially, an adept explainer forms assumptions (beliefs) about the explainee as a basis for making decisions on how to proceed with the explanation. Further, the explainer will update these beliefs during interaction based on the feedback received or the understanding gained about the listener's abilities or personality. In other words, the explainer builds a partner model (PM) and utilizes it for adaptive explanation generation. We ask how intelligent autonomous agents can be equipped with similar abilities to successfully explain a certain domain to a human user. To that end we address the following questions: Which information about the human explainee and the running interaction needs to be captured in a suitable PM? How can we formally represent and update the PM during an explanatory interaction? And how can we use the PM to steer the explainer's decision-making?

Previous research has largely focused on mapping user feedback to a certain level of understanding \citep{axelsson_using_2020} or determining the mental state of a listener \citep{buschmeier_communicative_2018}. These approaches do not encompass the full spectrum of adaptive processes observable in human interaction. In addition to tailoring explanations to the interlocutors' comprehension, we also need to consider user features that go beyond mere comprehension and can be gleaned from input received during the interaction, especially the user's feedback. Moreover, previous approaches to infer such features \citep{robrecht_snape_2023} use discrete and categorical values, ignoring the uncertainty in this process and how it can vary (decrease or increase) due to, e.g, unexpected user feedback.

In this paper we present SNAPE-PM, a formal approach to building a partner model (PM) and using it for adaptive explanation generation. We base this model on empirical findings showing that the users' domain expertise, cognitive load, attentiveness, and cooperativeness are most relevant for the underlying decisions. Regardless of whether such a feature is persistent (static) or floating (dynamic), our model employs Bayesian probabilistic inference to dynamically form and update potentially uncertain beliefs about them, based on the listener's feedback. We also address the challenge of realizing a decision process that is sufficiently flexible, precise and fast to select the next best explanatory actions given the current state of the PM. To meet all those requirements, the explanation generation is modeled as a non-stationary decision process shaped by the current PM.

In the remainder of the paper we first discuss related work, before we introduce the PM and its realization as a Dynamic Bayesian Network (DBN). Then we describe how explanation generation can be modeled as a non-stationary Markov Decision Process (MDP) rooted in the PM. Finally, we report results from an evaluation of SNAPE-PM, using simulations of different users with correspondingly differing consistent or changing feedback behavior.\footnote{Full platform independent python code is made available at \url{https://github.com/arobrecht/severus-study}}.

\section{Related Work}
Much work in the field of explainable artificial intelligence (XAI) has been directed to the automatic generation of single-turn explanations to make individual decisions of the system comprehensible. If XAI approaches adapt to the user's mental model, it only covers the estimated knowledge to detect differences to its own model \cite{sreedharan_foundations_2021, vasileiou_please_2023}.
\citet{miller_explanation_2019}, in contrast, suggests looking at human-human explanations to uncover what characterizes a good explanation. It quickly becomes evident that each explanation is an individual process, varying with the interlocutors' stance or goals or the domain they address \citep{keil_explanation_2006}. Involving the user as an active decision-maker in the modeling process is often called for but has rarely been implemented to date \citep{celikok_modeling_2023}. Especially everyday explanations show a high degree of variability \citep{fisher_exploring_2023}, which is not surprising if we see such an explanation as a process that is dialogically co-constructed by both interlocutors \citep{rohlfing_explanation_2021}. 

Following \citet{dillenbourg_symmetry_2016} and \citet{clark_referring_1986}, we conjecture that such a dialogue requires the explainer to have a suitable PM capturing the partner's understanding and other features. The term PM describes a dynamic and abstract model of the interlocutor as is needed to maintain a shared comprehension or grounding in a collaborative task \citep{dillenbourg_symmetry_2016}. A PM consists of \textit{dispositional} aspects (A's representation of B's long-term knowledge, skills, or traits) as well as \textit{situational} features (A's representation of B's current understanding, behavior, or intentions in the collaboration setting). Especially, dispositional aspects are often ignored in human-agent interaction approaches. Although they are rare \citep{anjomshoae_explainable_2019}, some approaches center their explanation around the user \citep{axelsson_you_2023, buschmeier_communicative_2018, idrizi_exploring_2024}. Most of such approaches focus on modeling user knowledge, although it is known that characteristics, experiences, expectations, and stereotypes significantly influence the global user model and, therefore, the interaction \citep{brennan_two_2010}.
Large Language Models (LLMs) represent a significant advancement in the field of dialog systems. When discussing adaptive explanations generated by LLMs, it's essential to note that mostly it is not the explanation itself that is adapted to the user; rather, it is the recommendation that has been optimized based on the user's profile. This recommendation is subsequently explained using an LLM \citep{lubos_llm_2024}. Furthermore, \cite{kunz_properties_2024} discuss GPT-4's ability to adapt to specific user groups, but do not clarify how the system identifies or categorizes users into these groups. In their research, \cite{macneil_experiences_2022} introduce an LLM generated code snippet together with different explanations. Those explanations are rated as useful by students. This shows the LLMs capabilities to generate different types of explanations, but there remains a lack of research on how these models determine which explanation type is most suitable for a particular user. Furthermore, \cite{kunz_properties_2024} discuss GPT-4's ability to adapt to specific user groups, but do not clarify how the system identifies or categorizes users into these groups. More focus must be placed on the initialization of the partner model.

Upon initializing a PM, it needs to be dynamically updated through verbal and non-verbal interaction \citep{dillenbourg_symmetry_2016}. One way to realize this constant update is to model the explanation as a dynamic process by using a hierarchical non-stationary MDP \citep{robrecht_snape_2023}, which has been shown to have beneficial effects on the user's understanding \citep{robrecht_study_2023}. 
MDPs are commonly used to model sequential decision processes and agent-environment interactions \citep{lapan_deep_2018}. Yet, they are rarely used to model verbal interactions such as explanations \citep{anjomshoae_explainable_2019}. Classically, an MDP \citep{puterman_markov_1994} is as a tuple $<S,A,T,R>$, consisting of a finite set of possible states $S$, a finite set of actions $A$, a transition model $T(s'|a,s)$ describing the probability to reach a state $s'$ after performing action $a$ in state $s$, and a reward model $R(a,s)$ defining the immediate reward that is obtained after performing action $a$ in state $s$. Alternative reward models used are $R(s|a,s')$ or $R(s)$.

MDPs typically assume the environment to be stationary, i.e the transition probability $T(s|a,s')$, set of action $A$, and definition of state $S$ to be constant throughout the decision process. However, in many real-life settings such as self-driving cars, stock market forecasting, or dialogue management, the environment may be non-stationary. Different approaches to deal with non-stationary MDPs (NSMDPs) have been proposed \citep{lecarpentier_non-stationary_2019}. One option is to model several experts or policies and to compare their predictions in order to minimize  \textit{regret} \citep{fruit_regret_2017}, an approach which is mainly used in competitive domains with, at least, one adversarial. Another approach -- called Hidden-Mode MDP (HM-MDP) \citep{hadoux_markovian_2015} -- is to break down the NSMDP into different modes, each of which having different transition or reward models. HM-MDPs have been used in several domains \citep{chades_momdps_2012} and are employed in the present work to adapt transition probabilities according to the current PM. Note that this is different from using so-called ``non-stationary policies'' \citep{scherrer_use_2012}, which can change for certain parts of a decision sequence and were proven to have the potential of being more efficient than classical policies.

As stated in \citep{luo_act_2024} the most prominent ways to solve an MDP are Reinforcement Learning and Monte Carlo Tree Search (MCTS). Although reinforcement learning is often used to train a near-optimal policy prior to applying it, MCTS can be used to solve the decision problem online. Current approaches combine both techniques to solve non-stationary environments using Policy-Augmented Monte Carlo tree search \citep{pettet_decision_2024}, while others decrease the size of the state space by breaking down the decision problem into sub-problems \citep{robrecht_snape_2023}. 

One of the biggest challenges when modeling adaptive explanations in human-agent interaction is for the agent to infer and update a sufficiently rich PM of the explainee. Analogous to previous related work by \citet{dillenbourg_symmetry_2016}, we aim at identifying the most relevant dimensions according to which the state of the PM can be specified and, importantly, can be estimated from information observable during the interaction. Going beyond the mere knowledge of the user (as the explainee), we opted for modeling four additional features that are inferred using a DBN and that influence the decision-making process governing the explanation behavior. As described next, this PM is pivotal for keeping track of and pursuing the main goal of the explanation: The grounding of the explanandum.

\section{The SNAPE-PM Model}
In this paper we introduce SNAPE-PM, a computational model for adaptive explanation generation. We first outline the overall structure of the model, to then describe in detail its main two processes: (1) The creation and update of the PM, and (2) the decision process that utilizes the PM for adaptive explanation generation.

Splitting such a cognitive process into sub-processes is a common procedure, e.g., in repair processes that are divided into inferring and repairing \citep{chandra_cooperative_2024}. In our model, the processes are carried out by different components. A Bayesian Network is used to analyze the users' feedback and map out the PM, while an MDP is solved via Monte Carlo Tree Search (MCTS) to determine the best next explanation move.
The sensitivity to changes in time, which is fundamental for an adaptive explanation, requires the Bayesian Network to be dynamic and the MDP to be non-stationary in order to allow the components to dynamically update one another. 
More specifically, as soon as the PM is updated due to received feedback from the explainee, a new MDP is constructed to formalize the decision problem arisen in the newly estimated explanation situation (with, e.g., new information needs or likely new effects of explanation moves). 

\begin{figure}[ht]
\centering
    \includegraphics[width=0.5\textwidth]{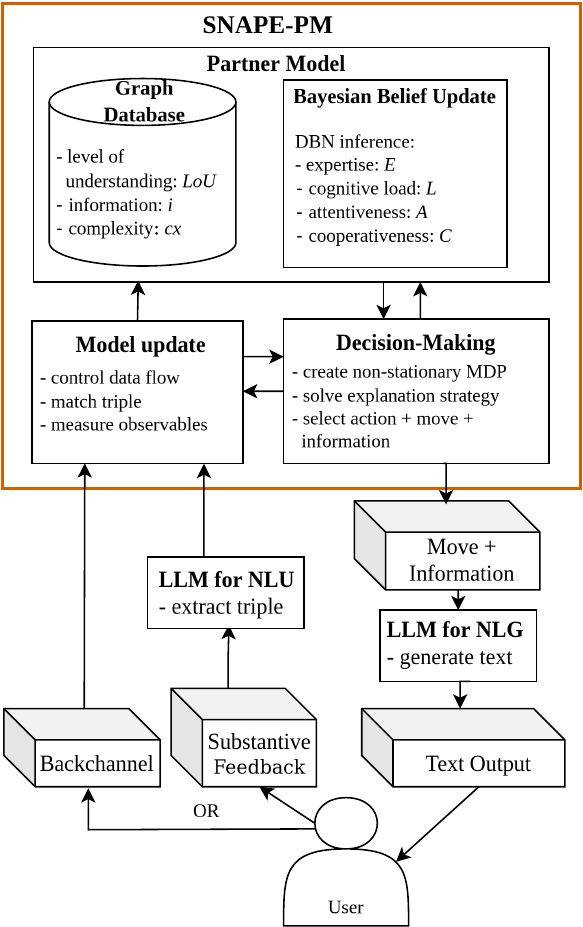}
    \caption{Overall architecture of the SNAPE-PM model.}
    \label{fig:1}
\end{figure}

At the behavioral level, one interaction cycle with the model always consists of potential user feedback and a corresponding explanatory utterance by the agent. This approach builds on the \textit{Sequential Non-stationary decision process model for AdaPtive Explanation} (SNAPE) proposed by \citep{robrecht_snape_2023}. We extend this approach by adding a comprehensive PM tailored for explanation generation (hence, SNAPE-PM), an inference mechanism that considers feature-wise interaction within the PM, a reformulation of the non-stationary MDP that rests on the PM, additional explanation moves, and a more complex reward function.
The general architecture of SNAPE-PM is shown in Fig. \ref{fig:1}.
The central step \textit{model update} manages the interplay between the partner model (PM) and the decision model. In addition, it controls the processing of user feedback and the flow of each agent's utterance. The PM comprises a graph-based knowledge base and a DBN inference model. The knowledge base (implemented using Neo4j \citep{inc_neo4j_2024}) contains the domain knowledge that can serve as explananda of differing complexity, and an estimate of the explainee's level of understanding $LoU$ of a specific information (triple in the knowledge graph). It is updated according to both, the user's feedback on the last presented information and the action(s) carried out by the agent. Positive feedback, regardless of its form (backchannel or substantive such as an affirmative question), increases the $LoU$ of the respective information while negative feedback (backchannel or questions expressing non-understanding) decreases it. Similarly, upon solving the MDP, the best next selected actions are handed back to \textit{model update}, which modifies the knowledge base accordingly and passes the actions to the Natural Language Generation (NLG) component. One of the main advantages of SNAPE-PM's modular structure is the generalizability of the system. Except for the graph database, none of the modules is domain-specific, but specific to interaction type of explanations.
The complete model has been implemented in an adaptive explainer agent for the scenario of explaining the board game Quarto. The clear separation of functionalities in combination with an abstract class formulation of the ontology management, partner model and model update enables an easy integration of a different ontology management system, partner model or decision process if wanted.

\subsection{Building the Partner Model}\label{sec:PM}
\citet{dillenbourg_symmetry_2016} describe a partner model as a "mosaic of fragments", underscoring the importance of different types of information about the partner. Consequently, it is essential to know which "fragments" are pertinent to the specific task at hand. For example, understanding the emotional state of the listener is crucial for motivational speeches or pep talks, while it assumes a diminished role in explanations. Similarly, comprehending the initial knowledge level of the recipient is important when explaining but not relevant when scheduling a meeting. 

In any case, a factored representation is needed to describe the state of a partner model. SNAPE-PM's partner model is realized as a DBN. A Bayesian Network (BN) is a graphical model representing a joint probability distribution in a factored form. DBNs are a specific kind of BNs, designed to model changes over time, assuming a stationary underlying process with the previous state as a prior \citep{murphy_dynamic_2002}. When building the DBN for representing a partner model in explanation, we take four factors into account: expertise, cognitive load, attentiveness, and cooperativeness. Each feature is dependent on its prior state, may be influenced by related other features, and can be tracked  (through filtering) when observing user feedback in the course of interaction.

The first feature to be considered in a partner model for adaptive explaining is the explainee's \textbf{expertise} $E$, as it influences the depth of information required for understanding.
We define the explainee's expertise as prior knowledge and experiences, differentiating it from the knowledge gained during the interaction, which is covered in detail in the graph-base knowledge model and referred to as understanding. Prior knowledge allows in-domain comparisons to related ideas and concepts; at the same time, it allows a deeper explanation by introducing additional information.

\textbf{Cognitive Load $L$} is the second considered feature and refers to a person's limited working memory resources used in a specific task \citep{chandler_cognitive_1991}. Including the user's cognitive load in the modeling of an interaction is relevant for the success of an adaptive system \citep{celikok_modeling_2023}. Previous studies have shown the effect of linguistic complexity on cognitive load \citep{engonopoulos_language_2013}. Therefore, we assume that adapting an explanation to the personal cognitive load is relevant to neither overload nor bore the user. Measuring the user's cognitive load is an established approach in HAI, and manifold linguistic measures are established to do so \citep{khawaja_measuring_2014, arvan_linguistic_2023}. The cognitive load is influenced by the level of expertise, as we expect a domain expert to have a higher capacity and a lower load in processing new but familiar information. Although the cognitive load is to a certain extent the opposite of expertise, it can also be influenced by other factors. Receiving a board game explanation, gaming experts can also have a high cognitive load if they are distracted by external factors, are tired or exhausted, or have general difficulties with verbal explanations. 

According to \citet{allwood_semantics_1992}, the communicative function of feedback can be illustrated as a ladder with four rungs: contact, perception, understanding, and attitudinal reactions. We consider all of these functions important for adapting explanations and thus the underlying PM. We thus introduce \textbf{attentiveness} $A$ to deal with the lower levels of feedback (contact and perception), and \textbf{cooperativeness} $C$ to represent the user's willingness to express understanding and attitude. Consequently, attentiveness reflects backchannels, while cooperativeness mainly considers substantive contributions where the user takes the turn \citep{chi_observing_2008}. We expect a user with high attentiveness to have a low probability of missing a given information; a cooperative user is assumed to autonomously interrupt and report non-understanding. Following the concept of downward evidence \citep{clark_using_1996} both features are dependent on another. When a user gives substantive feedback displaying (mis)understanding, successful contact and perception are also displayed.

\begin{figure}[ht]
    \centering
    \includegraphics[width=0.5\linewidth]{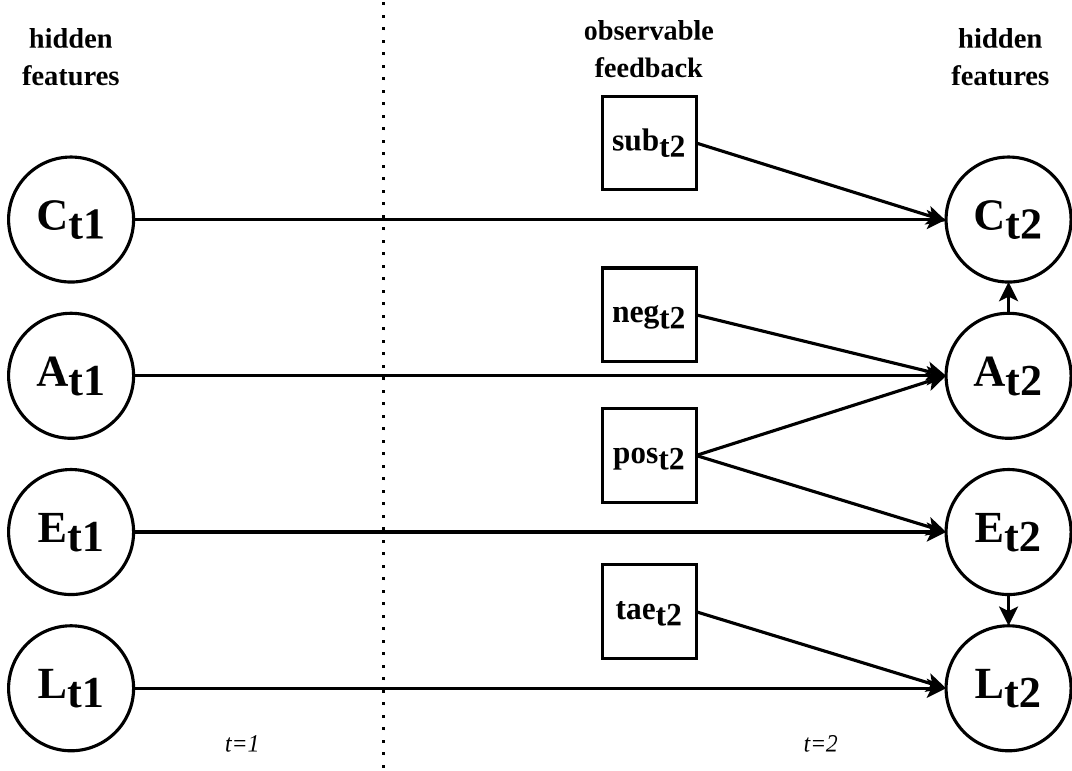}
    \caption{Dynamic Bayesian Network to build the partner model.}
    \label{fig:2}
\end{figure}
\begin{figure}
     \centering
     \begin{subfigure}[b]{0.45\textwidth}
         \centering
        \includegraphics[width=\linewidth]{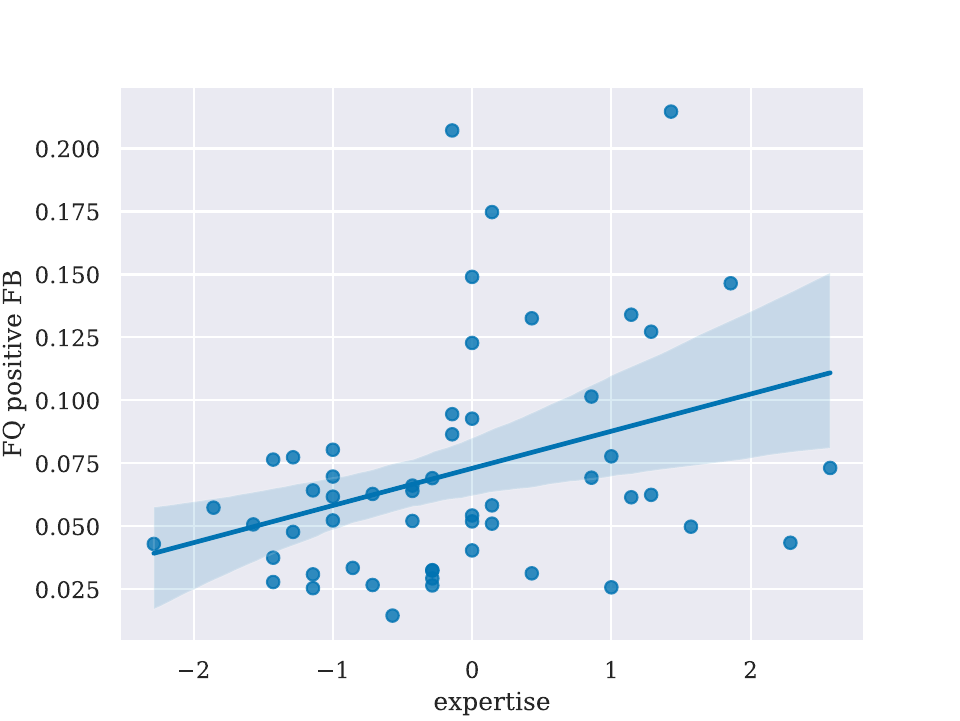}
         \caption{Expertise and positive backchannels}
         \label{fig:3a}
     \end{subfigure}
     \hfill
     \begin{subfigure}[b]{0.45\textwidth}
         \centering
         \includegraphics[width=\linewidth]{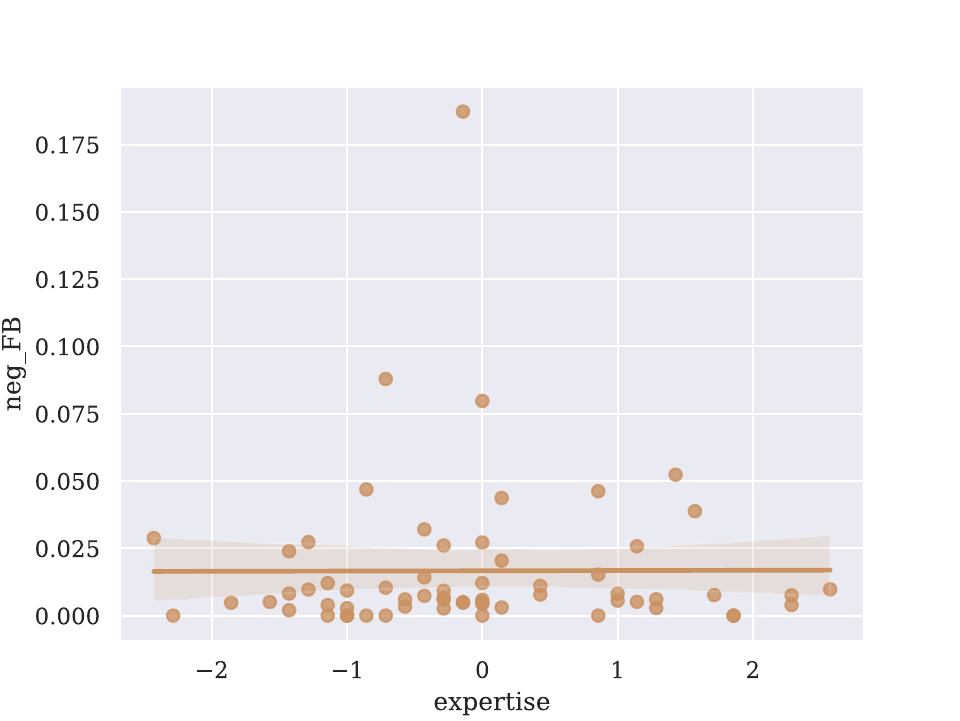}
         \caption{Expertise and negative backchannels}
         \label{fig:3b}
     \end{subfigure}
    \caption{Relation of expertise and feedback. \textbf{(a)} Linear regression of self-estimated level of expertise and frequency (FQ) of positive backchannel feedback (FB) provided by the participant \textbf{(b)} Linear regression of self-estimated level of expertise and frequency (FQ) of negative backchannel feedback (FB) provided by the participant}
     \label{fig:3}
\end{figure}

In order to measure the four PM features quantitatively and to formulate parameters to capture their dependencies and interrelations in a DBN (see  Fig.\ref{fig:2}), we rely on a variety of evidence and findings. Regarding expertise, we build on data from interaction studies carried out with the adaptive SNAPE model \citep{robrecht_study_2023}, which found a correlation between self-estimated level of expertise and the amount of positive feedback provided (Fig. \ref{fig:3a}). No correlation between negative feedback and expertise was found, see (Fig.\ref{fig:3b}). For simplicity, observable feedback is currently represented in terms of binary variables with value (\textit{yes}) or (\textit{no}). The formula to estimate the level of expertise based on the given feedback can be found in equation \ref{eq:expertise}.
\begin{eqnarray}\label{eq:expertise}
  P(E_{1:T},pos_{1:T}) = P(E_1)P(pos_1)\prod_{t=2}^T P(E_t|E_{t-1})P(pos_t|E_t)
\end{eqnarray}  
\begin{eqnarray}
  \text{with }p = \frac{positive}{utterances} \nonumber
\end{eqnarray}

Cognitive load (Eq. \ref{E:cognitiveload}) can be inferred from specific linguistic features in verbal interaction, such as word count \citep{khawaja_measuring_2014} (Eq. \ref{eq:WC}), Type-Token Ratio  \citep{arvan_linguistic_2023} (Eq. \ref{eq:TT}), or Gunning Fog Index \citep{gunning_technique_1968, khawaja_measuring_2014} (Eq. \ref{eq:GFI}). An example of how those measures are calculated for an utterance taken from the ADEX corpus \citep{fisher_adex_2024}.
\ex. German original and english translation taken from VP08\_ts837:
    \a. \textit{ja ich dachte vorher okay ich dachte vorher vorher hab ich's nicht verstanden. Weil ich dachte es gäb nur lange Steine und dann kurze und dann.}
    \b. \textit{yes I thought before okay I thought before before I did not understand. Because I thought it exists only long stones and then short and then.}

\begin{itemize}
    \item \textbf{Word Count:} The full amount of words in a text.
    \begin{eqnarray}\label{eq:WC}
      WC = \sum_{n=1}^{n} t_i = 26
    \end{eqnarray}  
    \item \textbf{Type-Token-Ratio:} The relation between the wordforms in a text (token) and the amount of basic lexems/words.
    \begin{eqnarray}\label{eq:TT}
      TTR = \frac{types}{token} = \frac{17}{26} = 0.6538
    \end{eqnarray} 
    \item \textbf{Gunning Fog Index:} The readablity of a text (17: college graduate, 12: high school senior, 6: sixth grade). A word is considered to be complex, if it consists of three or more syllables (suffixes are not included).
    \begin{eqnarray}\label{eq:GFI}
      GFI = 0.4[(\frac{words}{sentences})+100(\frac{complex}{words})] = \\ \nonumber 0.4[(\frac{26}{2})+100(\frac{1}{26})] = 9.4
    \end{eqnarray}
\end{itemize}
However, those measures are established for spoken interaction, while we restrict our current implementation of SNAPE-PM to typed user input, in which those measures are less indicative. When typing we rather delete than repeat; instead of stumbling, we pause. We thus turn to other observables such as mouse movement \citep{grimes_mind_2015} or typing speed and erasing behavior in relation to the person's mean. 
Specifically, we consider the time needed to type a sign ($t$) and the number of deleted signs ($d$) in the current feedback input and compare it to the overall mean of the current user ($x_{user}$) (Eq. \ref{E:ead}). If the current value is above the mean, the typing and erasing observable $tae$ increases; if it is below the current mean, it decreases (Eq. \ref{E:cognitiveload}). As shown in \citep{khawaja_measuring_2014} and discussed earlier, the calculation of load $L$ can not only be deduced from the typing behavior $tae$, but is also related to the expertise.
\begin{eqnarray}\label{E:cognitiveload}
  P(L_{1:T},tae_{1:T},E_{1:T}) = P(L_1)P(tae_1,E_1)\prod_{t=2}^T P(L_t|L_{t-1})P(tae_t,E_t|L_t)
\end{eqnarray}  

\begin{eqnarray} \label{E:ead}
    e = \begin{cases}
     higher, & \text{if } t>t_{user} \text{ and } d>d_{user}  \\
     lower, & \text{if } t<t_{user} \text{ and } d<d_{user} \\
     None, & \text{otherwise}
 \end{cases}
\end{eqnarray}
There are different ways to measure and infer attentiveness. The same listener has been shown to produce less feedback in identical tasks if distracted \citep{buschmeier_are_2011}. Therefore, the frequency of feedback can be taken as a marker for the user's current level of attentiveness (Eq. \ref{E:attentiveness}). Like the cognitive load, the user's attentiveness is dynamically changing while the explanation evolves. A low level of attentiveness leads to a higher probability of fully missing an utterance when utilizing this feature in the decision process (Sec. \ref{sec:MDP}).

\begin{eqnarray}\label{E:attentiveness}
  P(A_{1:T},pos_{1:T},neg_{1:T}) = P(A_1)P(pos_1,neg_1)\prod_{t=2}^T P(A_t|A_{t-1})P(pos_t,neg_t|A_t)
\end{eqnarray} 

The dynamic feature of cooperativeness can be measured implicitly through the amount of substantive feedback $s$ provided (Eq. \ref{E:cooperativeness}). In turn, a higher level of cooperativeness leads to a higher expected understanding when no feedback is provided (Sec. \ref{sec:MDP}).
\begin{eqnarray} \label{E:cooperativeness} 
  P(C_{1:T},sub_{1:T},A_{1:T}) = P(C_1)P(sub_1,A_1)\prod_{t=2}^T P(C_t|C_{t-1})P(sub_t,A_t|C_t)
\end{eqnarray}

Each of the four features is modeled as a latent variable node in the DBN and can have three values: $low$, $medium$, and $high$. The observables are binary variables (values $yes$ and $no$), except for $tae$ which has three values ($lower$, $higher$, $None$). All observables are directly gleaned from user feedback, which can be either positive or negative backchannels, or substantive feedback. Backchannels are provided by clicking on the matching smiley, or they can provide substantive feedback by using the open text option. As soon as the user clicks into the text box, the explanation is paused. A simplified and fully user-centered version of turn-taking. The presence or absence of feedback is taken as evidence and the DBN is started to update the agent's beliefs. It is implemented based on \citep{ducamp_agrumpyagrum_2020}, with the number of time steps set to 1000 as upper bound for the length of simulated conversations.

\subsection{Utilizing the Partner Model}\label{sec:MDP}
\begin{figure}[ht]
    \centering
    \includegraphics[width=0.7\textwidth]{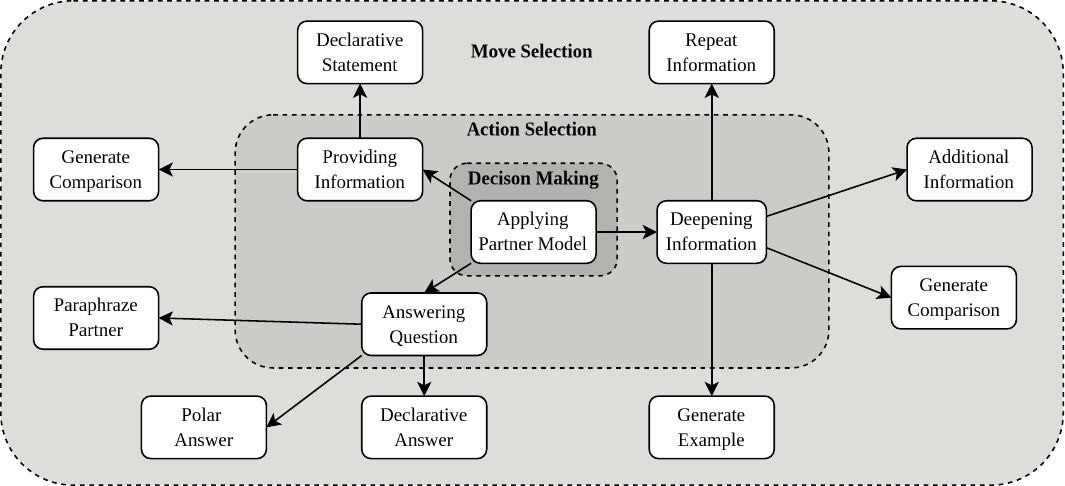}
    \caption{Available combinations of actions and moves.}
    \label{fig:4} 
\end{figure}

As the partner model is updated with each iteration, the decision-making component utilizing this model must also be dynamic. We thus model the decision process as a non-stationary MDP. To achieve real-time capability -- a key requirement for a language-based interactive system -- we break down the decision problem into semantically related explanation blocks that can be solved in realtime using MCTS. This approach reduces the search space and is based on insights into how humans structure their explanations \citep{fisher_exploring_2023}. The decision model decides which \textit{action} (what to say) and \textit{move} (how to say it) to perform next. Three different actions can be selected, each of which realized by two to four different moves (Fig. \ref{fig:4}), which differ at the rhetorical level and are based on typical explanation moves found in human-human explanations \citep{fisher_adex_2024}. To model the decision process of an adaptive explanation as a non-stationary MDP has been proposed before \cite{robrecht_snape_2023}. In contrast to the previous model, SNAPE-PM introduces some fundamental updates: The MDP (1) utilizes the DBN based PM for its decisions, (2) considers a bigger action space, and (3) uses preconditions and graph distances in the reward functions.

A state in the MDP is defined as:
\begin{itemize}
    \item \textbf{grounded} information in current block $b$: $G_b$
    \item \textbf{total} information in current block $b$: $T_b$
    \item active \textbf{question}: $q \in \{None, polar, open\}$ 
    \item features of the PM: $E,L,A,C$ 
    \item set of information that is \textbf{currently under discussion}: $CuD = \{i_1, ..., i_x\}$ and their levels of understanding ($LoU$)
\end{itemize}

Reward $r$, transition probabilities $t$, and level of understanding $LoU$ are calculated depending on the current state of the DBN. The reward of each action depends mainly on the predicted state of comprehension. An increase in this highly depends on the user's cognitive load and expertise, as we assume an expert with a low cognitive load to understand faster. The transition probability to stay in the previous state of the MDP, if no feedback is received, is higher as long as attentiveness or cooperativeness are estimated to be low, as the user is considered to have a high probability of being distracted. That is, generally, the user's knowledge and the explanation history affect which action is selected. The subsequent selection of explanation moves also depends on the PM features introduced before, as different moves have different success rates for different partners. In the following, we provide a detailed look at each action individually and describe how reward, level of understanding, and transition probability are defined for each action.

When \textbf{providing information}, a new aspect of the game is introduced by the explainer. The agent can provide information in two different ways: Information can be introduced using a declarative (\textit{Your opponents picks the next figure for you.}) or using a comparison (\textit{In contrast to chess you do not decide which figure to use next, your opponent does.}). Both options have benefits and drawbacks for different partners depending on their expertise. An expert is likely to benefit from comparisons to other board games, while a lay person is more likely to be confused by comparisons to unfamiliar games such that simple declarative statements are preferred. For this reason, the initial $LoU$ for a declarative increases linearly with growing expertise (Eq. \ref{E:LoU_declarative})\footnote{$\alpha$ is a value between 0 and 1, which is currently set to 0.5}, while the increase is exponential when using a comparison (Eq. \ref{E:LoU_comparison}). In result, an expert gains a higher initial $LoU$, but at a certain state of expertise the preferred method of providing information switches from declarative to comparison. The agent can also decide to introduce multiple information at once, depending on the triple complexity $cx \in [1,3]$ and the current user capacity $v$ (Eq.\ref{E:capacity}), which is dependent on the cognitive load and the value $\kappa$\footnote{$\kappa$ is currently set to 5 and can be increased for a more compact explanation}.
\begin{eqnarray} \label{E:LoU_declarative}
    LoU_{P-D}' = \left(1+\frac{\mathbb{E}(E)\cdot\alpha}{cx_i}\right)\cdot 0.5
\end{eqnarray}
\begin{eqnarray} \label{E:LoU_comparison}
    LoU_{P-C}' = \left(1+\frac{\mathbb{E}(E)^2}{cx_i}\right)\cdot 0.5
\end{eqnarray}
\begin{eqnarray}\label{E:capacity}
    v = (1- \mathbb{E}(L))\cdot\kappa)
\end{eqnarray}

Based on the current capacity, SNAPE-PM needs to balance the difficulty of the next utterance. $v$ should equal the combined complexity of the set of selected information $cx(i)$. For example: If the estimated cognitive load $\mathbb{E}(C)$ is 0.6, the capacity is two. The agent now looks for all information combinations that (1) are mandatory but not provided so far, (2) have a graph distance of one to ensure semantic connection, and (3) have a combined complexity equal to or closest to $v$. That is, a combination of either two information semantically connected units with the complexity of one (e.g. Quarto-has-strategy, strategy-is-complex) or a single utterance wit complexity two (passive-is-strategy) is suitable. In the context of this action, the transition probability (Eq. \ref{E:t_provide})\footnote{The transition probability to not reach $s'$, but stay in the previous state $s$ is always $1-t(s'|s,move)$} for both moves is equivalent, as an information has been said and cannot be introduced again.

\begin{eqnarray}\label{E:t_provide}
    t(s'|s,provide) = 1
\end{eqnarray}

While $LoU$ and $t$ need to be calculated at the move level, the reward considers the current state of knowledge and is calculated at action level. The reward for providing information (Eq. \ref{E:R_provide}) is defined as the sum of the average $LoU$ of each precondition $p$ of the current information $i$ minus one, as long as a non-introduced information exists; otherwise, the reward equals a highly negative value $\beta$. Further, although triples can have block-external preconditions, the reward function only considers internal ones as the agent cannot move to another block unless the current one is fully grounded or the user requests so.

\begin{eqnarray}\label{E:R_provide}
 R(s,provide) = \begin{cases}
     \frac{\sum_{p \in P} LoU_p}{|P_i|} - 1, & \text{if } \exists i \in T_b :LoU_i=None \\ 
     \beta, & otherwise
 \end{cases}   
\end{eqnarray} 

If an information has been introduced but remains ungrounded, due to no or negative user feedback, the agent has multiple options to \textbf{deepen information}: repeating what has been said before (\textit{It is not you who picks the piece for you, it is your opponent.}), giving additional helpful but not mandatory information (\textit{When you choose a piece for your opponent, you simply hand it to him.}), generating an example (\textit{for example, you could pick the big dark figure and pass it to your opponent.}), or drawing a comparison (\textit{In contrast to TicTacToe your opponent selects the figure for you.}). If a comparison or an example can be generated depends on the triple, if the triple allows a comparison depends on the distance of ontology's structure compared to (chess, TicTacToe, Bestof4, UNO) if an example is available is annotated in the ontology as well. When choosing which information to elaborate on, the information semantically closest to the last utterance $i_{t-1}$ is preferred. Including the graph distance $d$ here prevents the agent from jumping between information. Preconditions are not considered when deepening information, as the information has already been introduced, and graph distance takes precedence in the corresponding reward function (Eq. \ref{E:R_deepen}). 

\begin{eqnarray}\label{E:R_deepen}
 R(s,deepen) = \begin{cases}
     - d(i,i_{t-1}) -1, & \text{if } \exists i \in CuD :0 < LoU_i < g  \\
     \beta, & \text{otherwise}
 \end{cases}
\end{eqnarray} 

The increase in the estimated $LoU$ depends on the user's expertise $\mathbb{E}(E)$ and the complexity of the information $cx_i$. The moves repeat and example are well suited for lay explainees, as they do not require prior domain knowledge (Eq. \ref{E:LoU_deepen_RE}). The opposite holds for the moves additional information and comparison, which are suitable for expert explainees (Eq. \ref{E:LoU_deepen_AC}).

\begin{eqnarray}\label{E:LoU_deepen_RE}
LoU_{\text{rephrase/example}}' = LoU + \frac{1-LoU}{2} + \frac{\mathbb{E}(E)\cdot\alpha}{cx_i}
\end{eqnarray}
\begin{eqnarray}\label{E:LoU_deepen_AC}
LoU_{\text{additional/comparison}}' = LoU + \frac{1-LoU}{2} + \frac{\mathbb{E}(E)^2}{cx_i}
\end{eqnarray}

The moves example and comparison require the explainee to be highly attentive (Eq. \ref{E:t_deepen_EC}), while repeat and additional info are easier to perceive (Eq. \ref{E:t_deepen_RA}). Hence the MPD transition probability is directly influenced by the user's attentiveness.

\begin{eqnarray}\label{E:t_deepen_RA}
    t(s'|s,rephrase/additional) = \frac{1+\mathbb{E}(A)}{2}
\end{eqnarray}
\begin{eqnarray}\label{E:t_deepen_EC}
    t(s'|s,example/comparison) = \mathbb{E}(A)
\end{eqnarray}

In the current implementation of SNAPE-PM, the user can give substantive feedback by typing a question into a text box. If such a feedback occurs, it is handled by the \textbf{answering question} action. The previous state's $CuD$ is replaced by the requested information and $q$ is set to the according question type (\textit{polar} or \textit{open}). To reset the question value to \textit{None}, the question needs to be answered using the correct type of answer. The move selection, when answering a question, is influenced by the estimated cognitive load $\mathbb{E}(L)$, while the reward is influenced by the question value $q$ of the current state (Eq. \ref{E:r_answer}). If the user asked a question before, the reward is equal to zero; otherwise, it is equal to $\beta$ as it is a strong distortion of the dialog flow and causes confusion to answer a question that was not stated before.

\begin{eqnarray}\label{E:r_answer}
     R(s,answering) = \begin{cases}
     0, & \text{if } q \neq None \\
     \beta, & \text{otherwise}
 \end{cases} 
\end{eqnarray}

The content and type of a user question define content and type of the answer. A wh-question such as \textit{Where do I place the figure?} or a tag question such as \textit{What is the name again?} requires a declarative answer \textit{A-D} (\textit{You place it on the board./ The game is named Quarto.}), while a polar question (\textit{So it is a game for two?}), can either be answered with a polar answer \textit{A-P} (\textit{Yes.}) or by paraphrasing/summarizing the interlocutor \textit{A-S} (\textit{Indeed, a game for two.}). We include two moves \textit{polar answer} and \textit{summarize explainee} that can be used to answer the same type of question whilst being sensitive to the user's cognitive load $\mathbb{E}(L)$, which is expressed in the $LoU$ (Eq. \ref{E:LoU_polar}+\ref{E:LoU_summarize_ee}). The move \textit{declarative} addresses a different type of question (Eq. \ref{E:LoU_declarative_answer}). 

\begin{eqnarray}\label{E:LoU_polar}
    LoU'_\text{A-P} = \begin{cases}
        LoU + (1-LoU)\cdot(1-\mathbb{E}(L)), & \text{if } q = \text{polar}\\
        LoU, & \text{otherwise}
    \end{cases} 
\end{eqnarray}
\begin{eqnarray}\label{E:LoU_summarize_ee}
    LoU'_\text{A-S} = \begin{cases}
        LoU + (1-LoU)\cdot\mathbb{E}(L), & \text{if } q = \text{polar}\\
        LoU, & \text{otherwise}
    \end{cases}  
\end{eqnarray}
\begin{eqnarray}\label{E:LoU_declarative_answer}
    LoU'_\text{A-D} = \begin{cases}
        LoU + (1-LoU)\cdot\alpha, & \text{if } q = \text{open}\\
        LoU, & \text{otherwise}
    \end{cases}  
\end{eqnarray}
\begin{eqnarray}\label{E:tanswer}
    t(s|s',answering) = \begin{cases}
        1, & \text{if } i_a = CuD\\
        0, & \text{otherwise}
    \end{cases}
\end{eqnarray}

Note that the transition probability is always 1 as long requested information is addressed in the answer, as we consider an explainee who just asked a question to be attentive (Eq. \ref{E:tanswer}). This holds for all three moves. In addition to the previously discussed actions, the agent may also use a discourse structuring action comprising four different moves (mentalize partner, comprehension question, bridging block, summarizing block) to lead from the current block to a new one. This action will be included in the MDP in future work. 

As real-time processing is key for the system, we reduce the size of the MDP to include only a preselection of conversationally valid moves. That is, the MDP does not consider answer moves if no question was asked and does not look at provide moves if the appropriate triple was already provided. This restriction of the MDP enables real-time processing while still retaining choice over valid moves. Likewise, the MDP does not first choose the action and afterwards the move, but chooses from among all action and move combinations that are conversationally valid. 
For solving the decision process, Monte Carlo Tree Search (using the Python mcts library \citep{sinclair_mcts_2019}) is adapted to output the two best results. If these are both of the same action type and have a distance of one or less, both are presented to the user at the same time as a combined explanation. Note that the reward of a terminal state is equal to the cumulated reward of all actions, which in turn depend on the current PM as well as external factors such as the $CuD$ or $LoU$.

\section{Evaluation}
\begin{figure}
     \centering
     \begin{subfigure}[b]{0.45\textwidth}
         \centering
         \includegraphics[width=\linewidth]{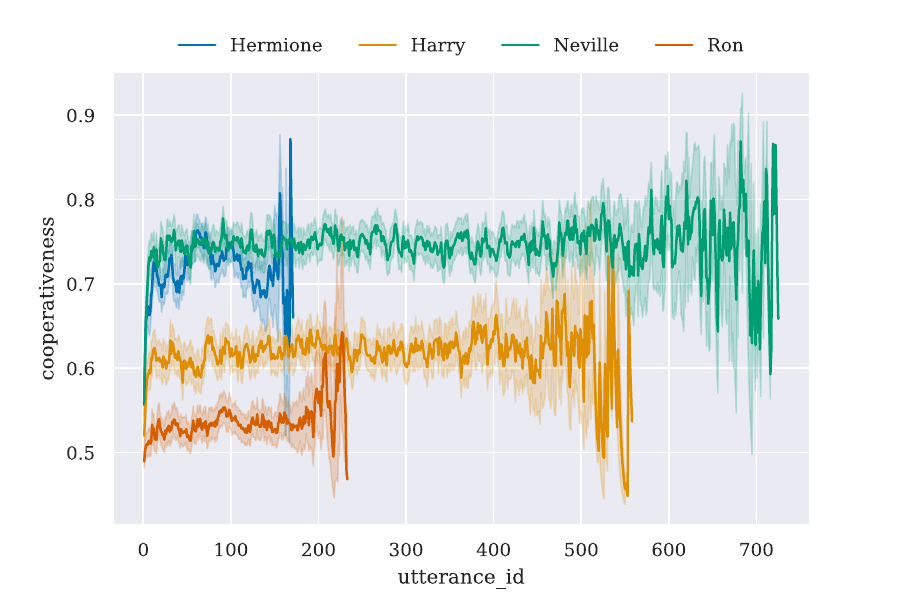}
         \caption{Expertise}
         \label{fig:6a}
     \end{subfigure}
     \hfill
     \begin{subfigure}[b]{0.45\textwidth}
         \centering
          \includegraphics[width=\linewidth]{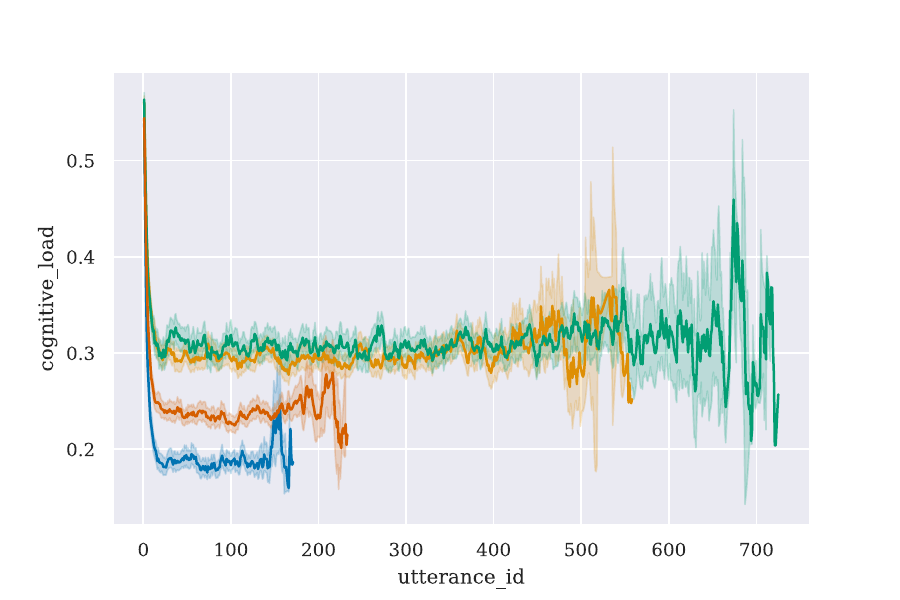}
         \caption{Cognitive load}
         \label{fig:6b}
     \end{subfigure}
     \hfill
     \begin{subfigure}[b]{0.45\textwidth}
         \centering
          \includegraphics[width=\linewidth]{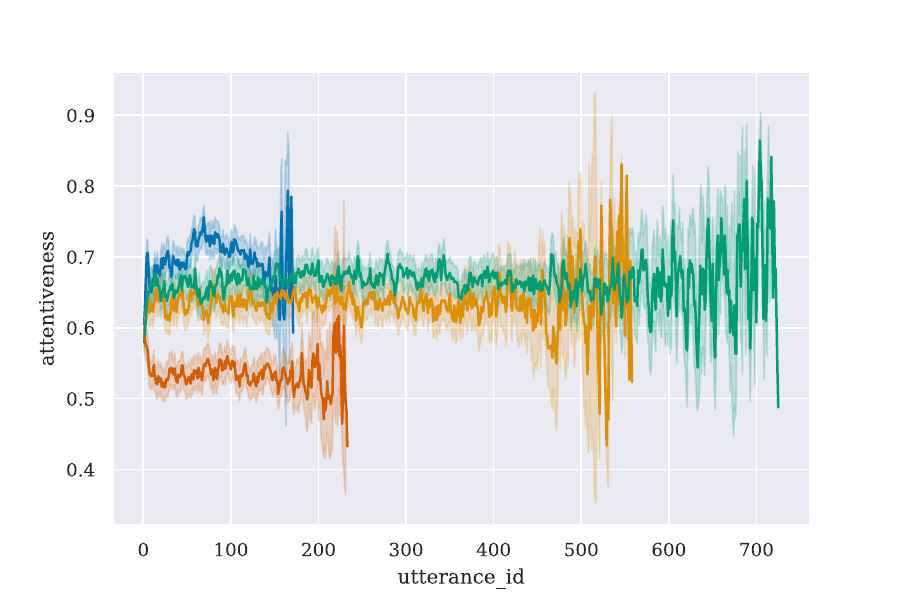}
         \caption{Attentiveness}
         \label{fig:6c}
     \end{subfigure}
     \hfill
     \begin{subfigure}[b]{0.45\textwidth}
         \centering
          \includegraphics[width=\linewidth]{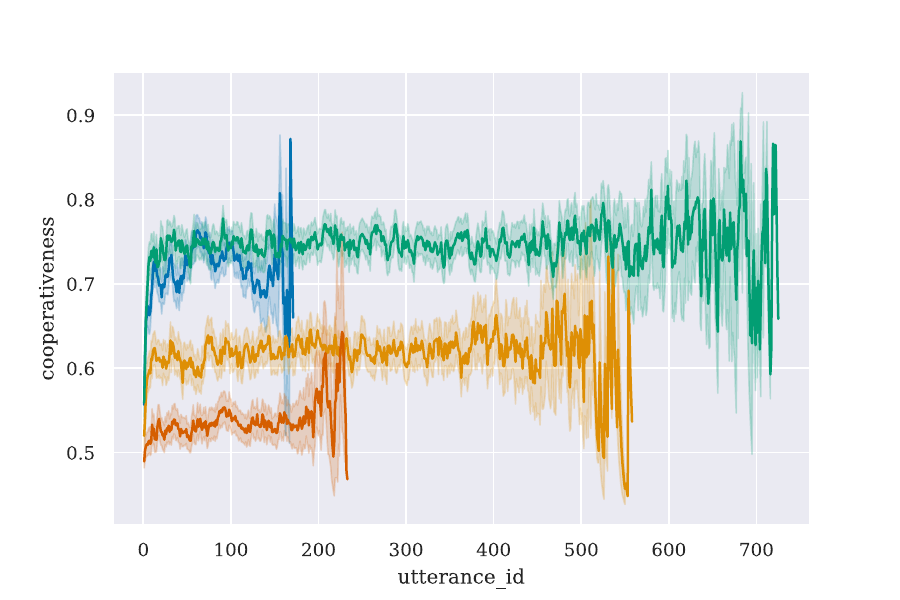}
         \caption{Cooperativeness}
         \label{fig:6d}
     \end{subfigure}
        \caption{Development of \textbf{(a)} expertise \textbf{(b)} Cognitive Load \textbf{(c)} Attentiveness \textbf{(d)} Cooperativeness. When simulating interactions between SNAPE and four different Personas.}
        \label{fig:6}
\end{figure}

To evaluate SNAPE-PM we investigated whether the model can autonomously adapt its strategies when interacting with different kinds of explainees. To do this, 140 interactions with each of five different partners generating consistent or even inconsistent feedback behavior were simulated.
\begin{table}[ht]
\caption{Probability Distributions for Automated Feedback Generation in Evaluation Personas as in feedback\_generation.py}
    \centering
    \begin{tabular}{cccccc}
    \toprule
         & \textbf{$P(no)$} & $P(bc)$ & $P(s)$ & $P(p)$ &$P(n)$  \\
    \midrule
        \textbf{Hermione} & 0.1 & 0.5 & 0.4 & 0.9 & 0.1\\
        \textbf{Harry} & 0.4 & 0.4 & 0.2 & 0.3 & 0.7\\
        \textbf{Ron} & 0.6 & 0.3 & 0.1 & 0.8 & 0.2\\
        \textbf{Neville} & 0.2 & 0.4 & 0.4 & 0.3 & 0.7\\
    \bottomrule
    \end{tabular}
    \label{tab:1}
\end{table}
The probability distribution over the different types of feedback -- no feedback ($no$), backchannel ($bc$) or substantive ($s$) -- and the tendency of feedback -- positive ($p$) or negative ($n$) -- for the different personas is shown in Table \ref{tab:1}. \textit{Hermione} and \textit{Neville} both want to learn and provide a lot of feedback. While Hermione prefers (mainly positive) backchannels over questions, as she is smart and has a lot of background knowledge, Neville has a hard time learning new things and asks more questions. \textit{Harry} and \textit{Ron} are less cooperative and do not produce feedback in 40\% of the time. While Ron loves to play chess and respectively gives very little negative feedback, as he is an expert in the board game domain, Harry is unfamiliar with games and rather provides negative feedback. 
\begin{figure}[t]
\centering
        \includegraphics[width=0.7\linewidth]{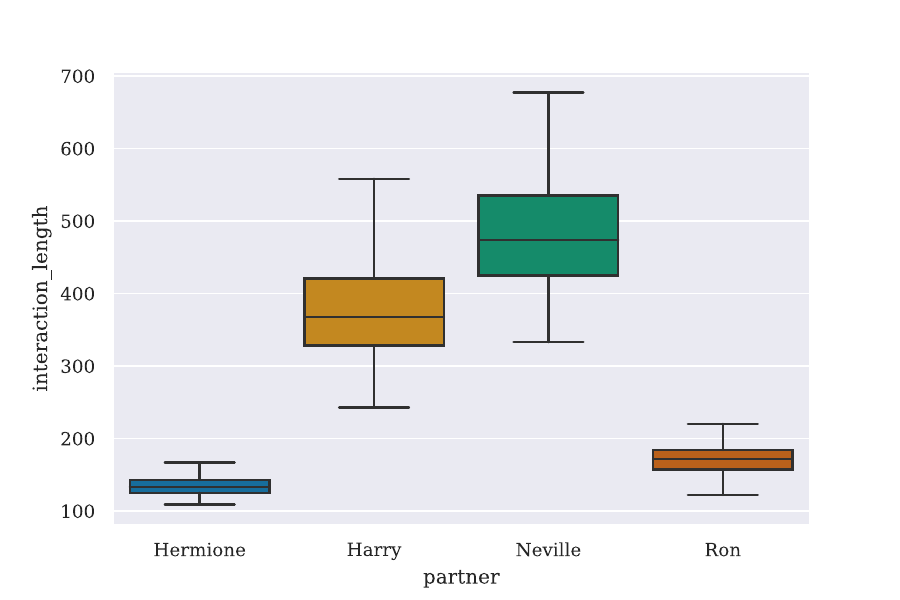}
        \caption{Length of explanation in iterations}
        \label{fig:5}
\end{figure}
When comparing the number of iteration cycles needed to generate a full explanation, we see the high impact of expertise and attentiveness on the length of the interaction (Fig. \ref{fig:5}). While SNAPE-PM considers Hermione to understand the fastest (m=134.764; std=12.654), Nevilles explanations are the longest and are subject to the strongest fluctuations, which also leads to the highest standard deviation (m=481.829, std=83.867). In terms of interaction length, Harry and Ron are in the middle range, with an explanation for Ron being quicker (m=173.079; std=22.292) than for Harry (m=375.007; std=65.890).
When looking at the trajectories of the partner model features, we see different values for the different partners. Although Harry ($m=0.474; std=016$) and Neville ($m=0.471; std=0.161$) both have an expertise slightly below 0.5, Ron's expertise is in the middle ($m=0.614; std=0.163$) and sometimes drops below 0.5, and Hermione's ($m=0.786; std=0.129$) expertise is clearly and constantly above 0.5 (Fig. \ref{fig:6a}). The cognitive load (Fig. \ref{fig:6b}) is clearly below 0.3 for all fours partners (Harry: $m=0.3; std=0.0719$, Hermine: $m=0.197; std=0.072$, Neville: $m=0.31; std=0.085$, Ron: $m=0.245; std=0.066$). The attentiveness is above 0.5 for all partners, led by Hermione ($m=0.701; std=0.123$), followed by Neville ($m=0.666; std=0.131$), Harry ($m=0.635; std=0.136$) and Ron ($m=0.536; std=0.132$) (Fig. \ref{fig:6c}). Cooperativeness is led by Neville ($m=0.7476 std=0.115$), followed by Hermine ($m=0.718; std=0.124$), Harry ($m=0.619; std=0.123$) and Ron ($m=0.532; std=0.105$) (Fig. \ref{fig:6d}).

\begin{figure}
     \centering
     \begin{subfigure}[b]{0.45\textwidth}
         \centering
         \includegraphics[width=\linewidth]{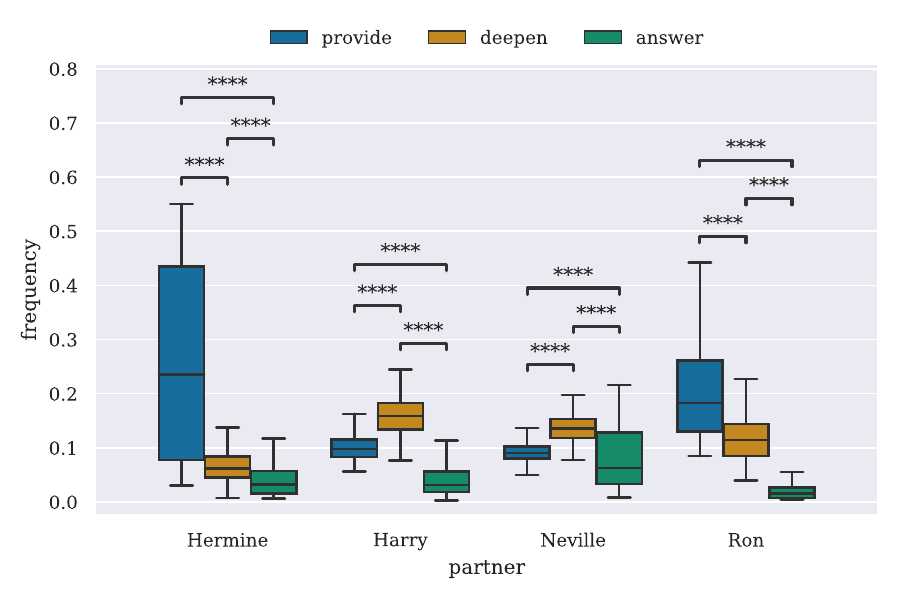}
        \caption{All actions}
        \label{fig:7a}
     \end{subfigure}
     \hfill
     \begin{subfigure}[b]{0.45\textwidth}
         \centering
         \includegraphics[width=\textwidth]{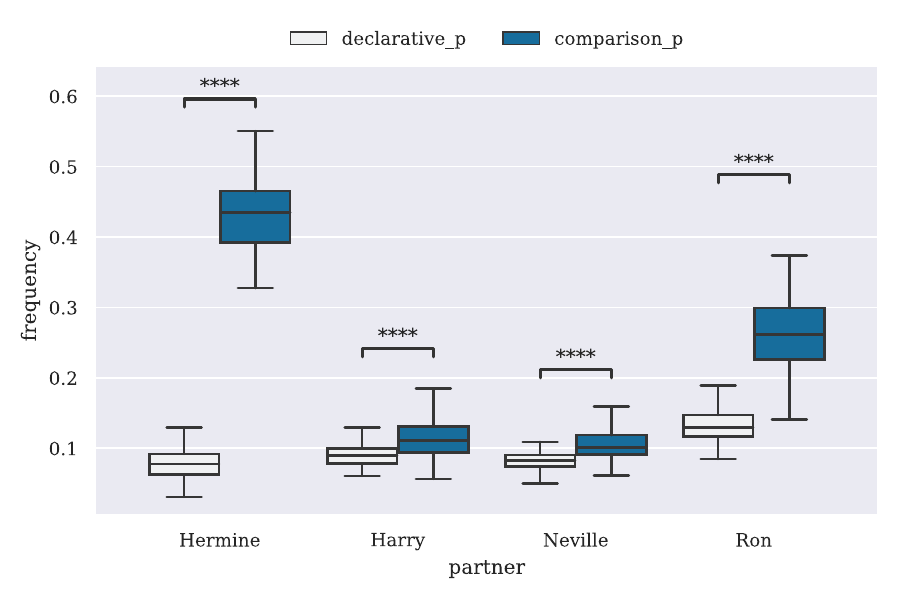}
        \caption{Providing action}
        \label{fig:7b}
     \end{subfigure}
     \hfill
     \begin{subfigure}[b]{0.45\textwidth}
         \centering
         \includegraphics[width=\textwidth]{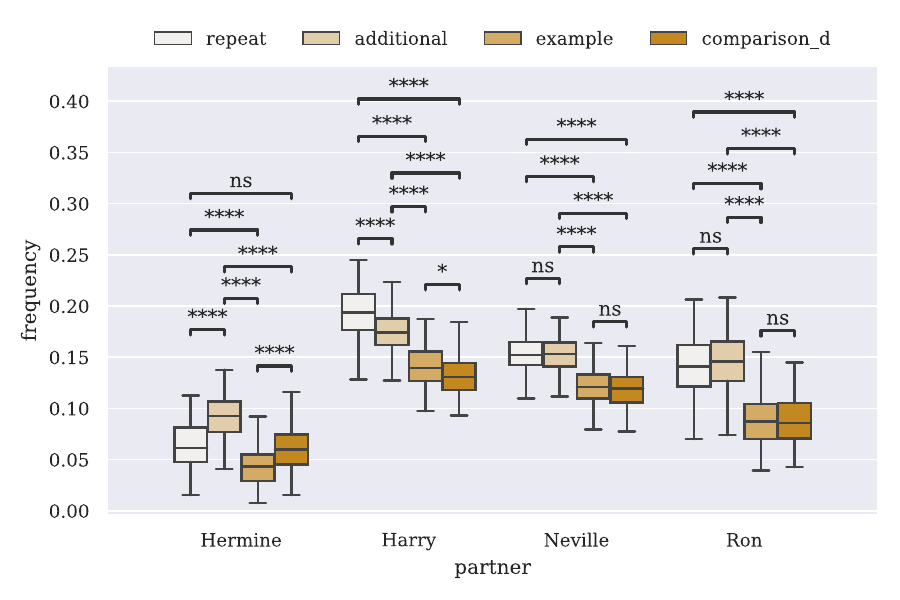}
        \caption{Deepening action}
        \label{fig:7c}
     \end{subfigure}
     \hfill
     \begin{subfigure}[b]{0.45\textwidth}
         \centering
         \includegraphics[width=\textwidth]{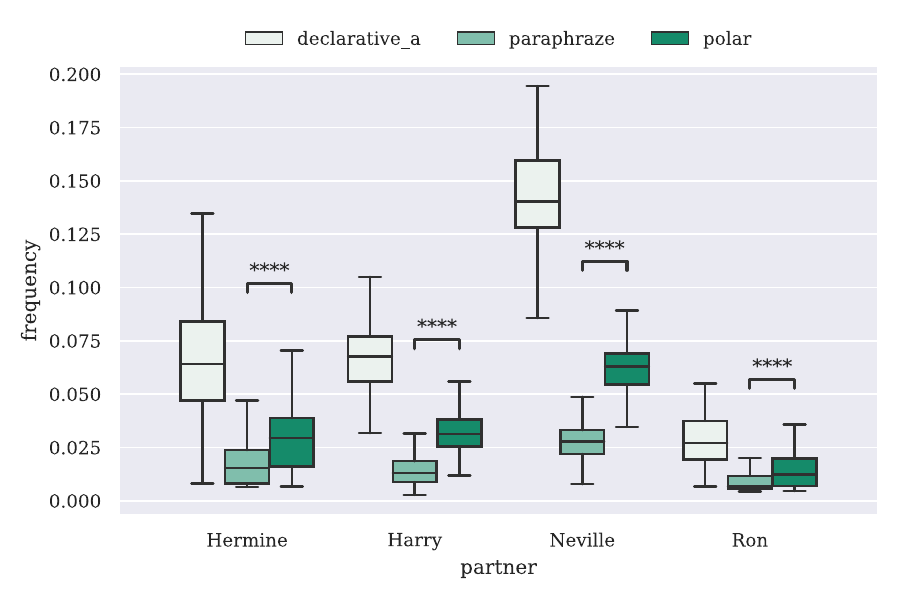}
        \caption{Answering action}
        \label{fig:7d}
     \end{subfigure}
        \caption{Distribution of \textbf{(a)} explanatory actions and \textbf{(b)} moves produced for the provide action, \textbf{(c)} moves produces for the deepening action, \textbf{(d)} moves produced for the answering action. produced by SNAPE-PM with different explainee personas.}
    \label{fig:7}
\end{figure}

As shown in Figure \ref{fig:7a}, an independent t-test with Bonferroni correction\footnote{p-value annotation legend:
$ns: 5.00e^-02 < p <= 1.00e+00; *: 1.00e-02 < p <= 5.00e-02; **: 1.00e-03 < p <= 1.00e-02; ***: 1.00e-04 < p <= 1.00e-03; ****: p <= 1.00e-04$} shows that the actions selected by SNAPE-PM highly depend on the partner the system is interacting with. While for Hermione ($p=1.012e-98; s=2.441e+01$) and Ron ($3.347e-76; s=2.071e+01$) the \emph{provide} is by far the most frequent action, SNAPE-PM prefers to use deepening moves more often for Harry ($9.991e-111; s=-2.633e+01$) and Neville ($p=1.923e-106; s=-2.564e+01$). This ratio can be explained primarily by the expertise of the individual partners.
The amount of answers is fully dependent on the amount of questions asked by the partner, which explains that there are more answers generated for Neville than for Ron, who has a low cooperativeness. Interestingly, the selection of moves for the different partners shows the interplay of the different features in the partner model. When providing information, SNAPE-PM has the choice between a declarative statement or a comparison (Fig. \ref{fig:7b}). This move selection only depends on the feature of expertise as the transition probability is always one, regardless of attentiveness or cooperativeness. 
Again an independent t-test with Bonferroni correction shows that, even though the comparison is favored over a declarative statement for each partner, the difference increases with an increase in expertise (p-values: Hermione $p=8.902e-192; s=-7.893e+01$, Ron $9.967e-80; s=-2.720e+01$, Neville $p=7.252e-22; s=-1.066e+01$, Harry $p=5.926e-17; s=-9.127e+00$). 
The second type of action is \emph{deepening information} (Fig. \ref{fig:7c}). Here, move selection considers the expertise and the attentiveness of the user, which makes the interpretation of the results more difficult, but also even more interesting. According to an independent t-test with Bonferroni correction following significant ANOVAS, \emph{additional} is preferred over \emph{repeat} and \emph{comparison} over \emph{example} for attentive domain experts, and the other way around for lazy non-experts. This becomes particularly clear when comparing \emph{repeat} and \emph{additional} for the partners Hermione ($p=1.319e-19; s=-1.021e+01$) and Harry ($1.563e-08; s=6.402e+00$). The second effect that can be observed concerns the use of complex moves compared to less complex moves. SNAPE-PM uses the moves \emph{repeat} and \emph{additional} with a higher frequency than the moves \emph{example} and \emph{comparison} for all partners, which can be explained with a higher probability to succeed. The agent always wants to increase the $LoU$ to a certain degree, yet should always choose the less complex utterance for this. We see a decrease of this preference with increasing attentiveness. Therefore, the difference between \emph{additional} and \emph{comparison} is much larger for Ron ($1.811e-46; s=1.780e+01$) than it is for Hermione ($p=2.615e-22; s=1.103e+01$). 
The difference between the usage of less complex moves (\emph{repeat, additional}) and more complex ones (\emph{example, comparison}) is highly significant for all partners, with one exception for Hermioneexcept for Hermione for whom the difference between \emph{repeat} and \emph{comparison} is not significant ($p=1.000e+00; s=9.417e-01$). 
This can be explained by a combination of the effects reported before.
In particular, \emph{comparison} is more likely to be chosen than \emph{example}, which is consistent with the explainee's high expertise.
The last action that needs to be considered in more detail is the \emph{answering} of stated questions (Fig. \ref{fig:7d}). The comparison between the declarative and the other two answers is not relevant, as they are clearly determined by the nature of the question and are mutually exclusive. When summarizing with polar answers, all partners prefer the polar answer. This can be explained by the partner's estimated cognitive load.

The fifth simulated person \textit{Luna} produced inconsistent feedback by altering her behavior every 30 steps: She can be attentive and clever (A-C), inattentive and clever (I-C), attentive and slow (A-S) and inattentive and slow (I-S) (see Tab.\ref{tab:2}).
\begin{table}[ht]
\caption{Probability Distributions for Automated Feedback Generation for Lunas Moods in feedback\_generation.py}
    \centering
    \begin{tabular}{cccccc}
    \toprule
    &\textbf{$P(no)$} & $P(bc)$ & $P(s)$ & $P(p)$ &$P(n)$  \\
    \midrule
        \textbf{A-C} & 0.1 & 0.6 & 0.3 & 0.7 & 0.3\\
        \textbf{A-S} & 0.1 & 0.6 & 0.3 & 0.3 & 0.7\\
        \textbf{I-S} & 0.7 & 0.2 & 0.1 & 0.3 & 0.7\\
        \textbf{I-C} & 0.7 & 0.2 & 0.1 & 0.7 & 0.3\\
    \bottomrule
    \end{tabular}
    \label{tab:2}
\end{table}

\begin{figure}[h]
    \centering
    \includegraphics[width=0.7\linewidth]{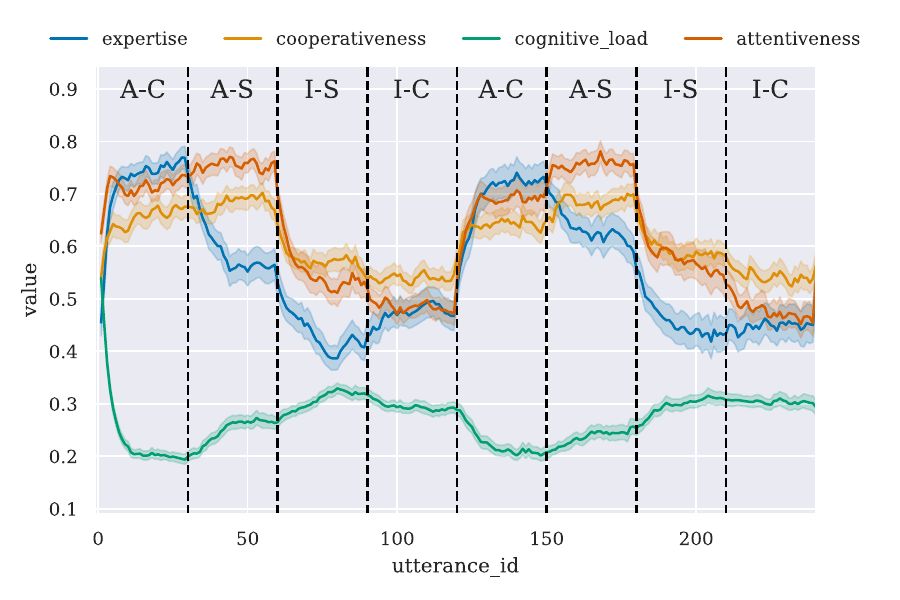}
    \caption{Changing PM states formed for an explainee who provides varying feedback behavior}
    \label{fig:8}
\end{figure}

Figure \ref{fig:8} shows the continuous adaptations that SNAPE undergoes during the first 250 iterations. The influence of the amount of feedback on the speed of adaptation should be particularly emphasized. For example, we see a rapid drop in expertise when switching from A-C to A-S, but only a very slow increase when switching from I-S to I-C. When switching to more feedback again (A-C) also the expertise rises fast again. This implicit connection between attentiveness and expertise can be explained by both values being dependent on the observed positive feedback.

\section{Conclusion}
This paper focuses on how partner models can be built and utilized to enable adaptive explanation in human-agent interactions. We presented SNAPE-PM, a formal model for representing and inferring relevant features of the explainee (using a DBN), and for turning this information into a dynamically adapted decision process for choosing the best next explanatory moves (by formulating and solving a non-stationary MDP). An implementation of this model was tested in simulations with five different personas, four of them showing different but consistent and one showing changing feedback behavior. The results demonstrate that SNAPE-PM adapts its explanations to different personas, and does so in a fast and continuous manner. This displays that the agent is able to build, update and utilize a partner model for adaptive explanation generation, especially leveraging on the features expertise, attentiveness and cooperativeness. 

\begin{figure}
     \centering
     \begin{subfigure}[b]{0.45\textwidth}
         \centering
         \includegraphics[width=\linewidth]{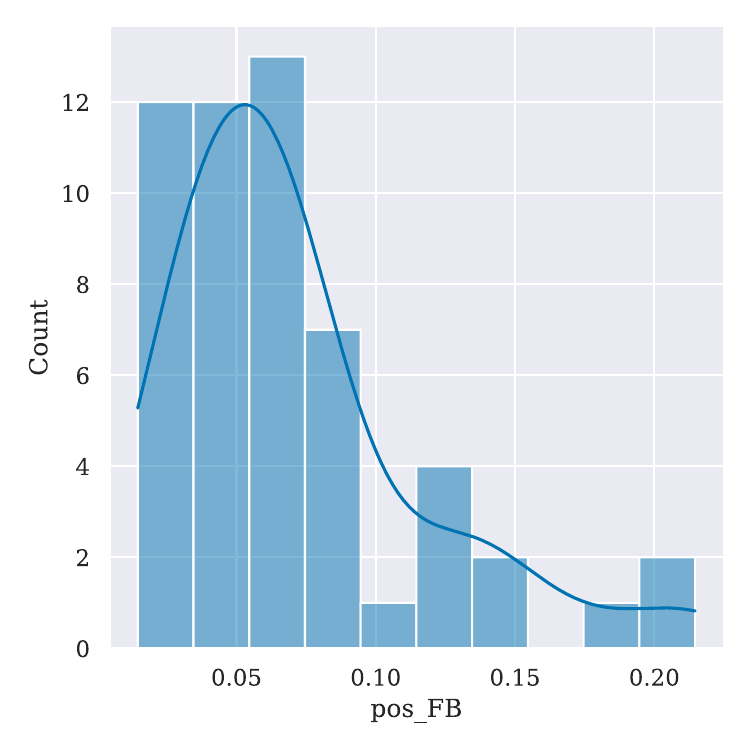}
         \caption{Positive feedback}
         \label{fig:9a}
     \end{subfigure}
     \hfill
     \begin{subfigure}[b]{0.45\textwidth}
         \centering
         \includegraphics[width=\linewidth]{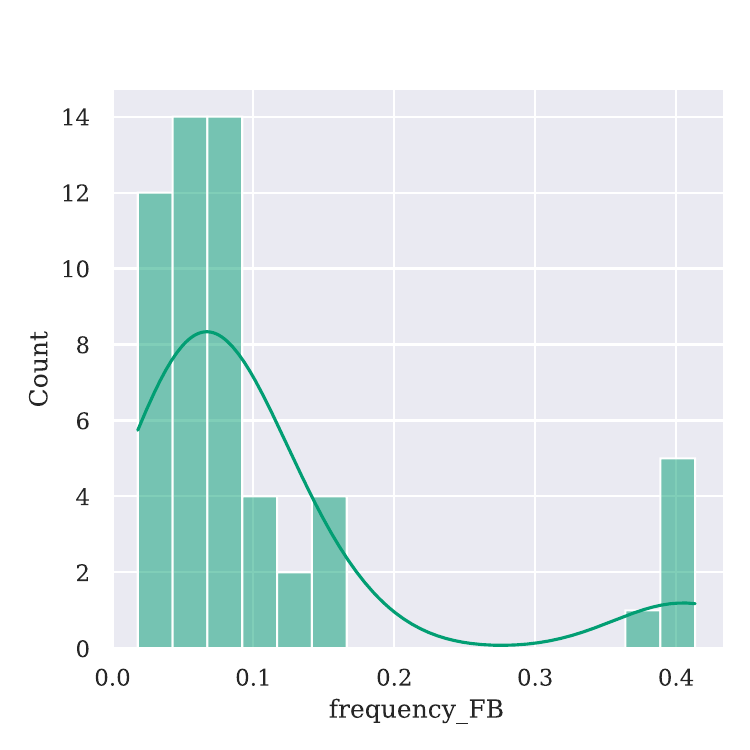}
         \caption{Backchannel-like feedback}
         \label{fig:9b}
     \end{subfigure}
    \caption{Frequency of \textbf{(a)} positive feedback \textbf{(b)} backchannel-like feedback; in human-agent explanations.}
    \label{fig:9}
\end{figure}

Clearly, our evaluation of the model has limitations. The restriction to simulated users with distinct types of feedback is a severe limitation as humans will behave differently and less systematic than our simulated personas. Further studies need to prove that real human-agent interactions are comparable to the simulated partners. Indeed, data from a related human-agent interaction study using another agent have shown a relatively low frequency of positive feedback and feedback in general (Fig. \ref{fig:9a}+\ref{fig:9b}) \citep{robrecht_study_2023}.

\begin{figure}
     \centering
     \begin{subfigure}[b]{0.45\textwidth}
         \centering
         \includegraphics[width=\linewidth]{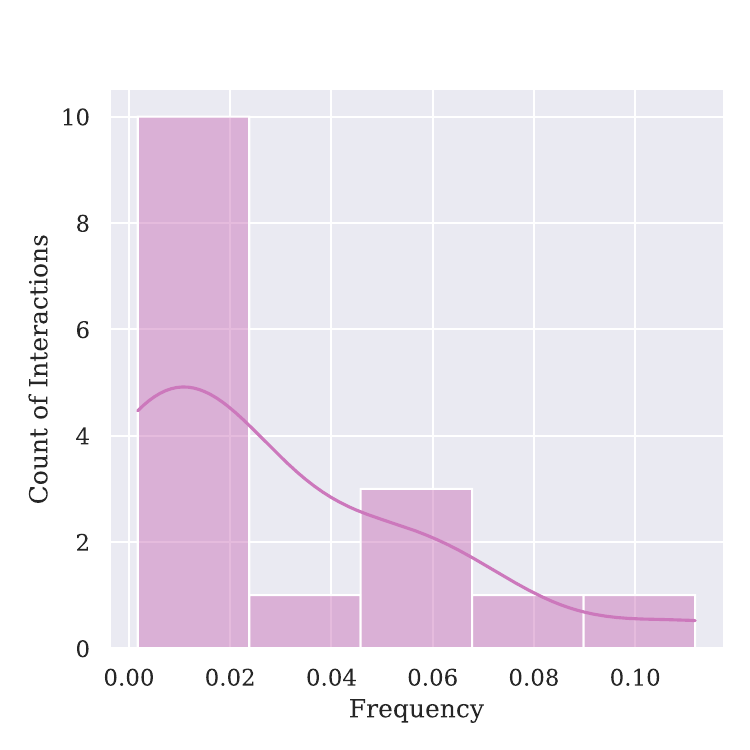}
        \caption{human-agent}
        \label{fig:10a}
     \end{subfigure}
     \hfill
     \begin{subfigure}[b]{0.45\textwidth}
         \centering
         \includegraphics[width=\linewidth]{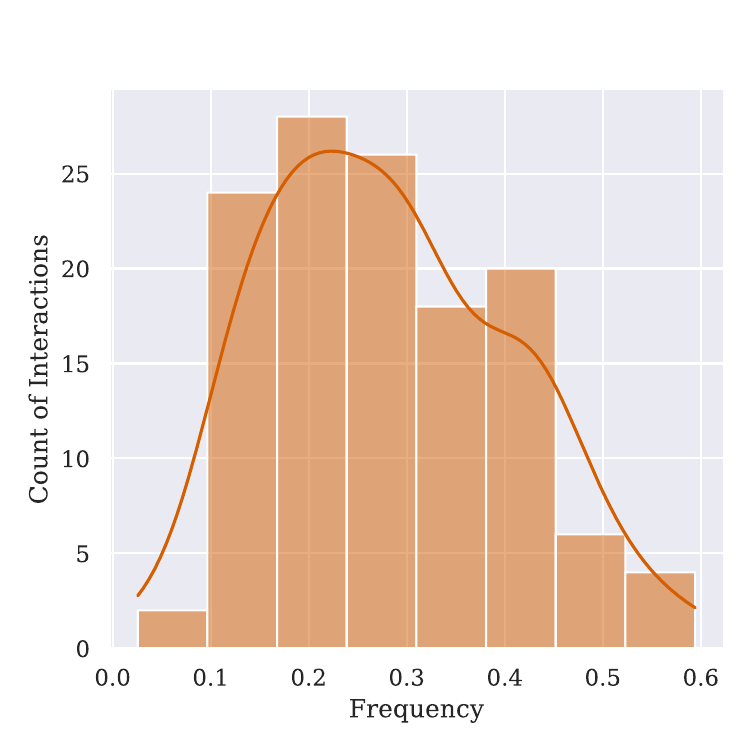}
        \caption{human-human}
        \label{fig:10b}
     \end{subfigure}
    \caption{Frequency of substantive feedback in \textbf{(a)} human-agent and {(b)} human-human interaction.}
    \label{fig:10}
\end{figure}

Likewise, the relative frequency of substantive feedback in (small-scale) human-human explanation studies was found to be nearly normally distributed over a range of 0 - 0.6, while in human-agent interaction this frequency was significantly lower (0.0 - 0.1) (Fig. \ref{fig:10a}+\ref{fig:10b}).
Notwithstanding these limitations, our simulation-based evaluation does provide a first proof of concept and allows for a focused view at individual strategies as well as a direct analysis of the viability of the underlying inference and decision-making mechanisms. Another limitations of the evaluation is that observables for cognitive load were equally distributed for each partner, and were only partner-dependent due to its relation to expertise. 

An important next step will be to equip SNAPE-PM with full-blown NLU and NLG components. Current LLM-based implementations are already being integrated and the extended SNAPE-PM model will then be tested in an online user study. This will allow for estimating and updating cognitive load from linguistic input provided by real users. Other future work will be directed to gathering user data about feedback frequencies in interactions with SNAPE-PM and adjusting (learning) parameters of the DBN accordingly. Finally, the model proposed here only scratches the surface of how explainers and explainees collaborate to co-construct an explanation. For example, it is common for explainees to take a more active role in the dialog by giving explicit feedback such as \textit{I often play board games.}, which should increase the estimated level of expertise, or \textit{Slow down, I think I lost you.}, a strong indicator of high cognitive load. The other way around, the explainer also should be able to directly query about understanding, expertise, attentiveness, cooperativeness or cognitive load of the explainee. Those complex yet natural forms of dialogical interaction are important means of propelling an explanation forward, but still pose important challenges for the field of XAI and human-agent interaction alike. The approach presented here to intertwine formal representational, inference, and decision-making models specifically for adaptive explanation generation may provide a principled basis for tackling those larger challenges and thus pave the way for making autonomous agents more explainable.

\bibliographystyle{unsrtnat}
\bibliography{references}  






\end{document}